\documentclass[letterpaper, 10 pt, conference]{ieeeconf}  %
\IEEEoverridecommandlockouts                              %
\usepackage{fp}
\FPset{\pb}{0}
\newcommand{\pagebudget}[1]{}
\newcommand{\showtotalpagebudget}[1]{}
\usepackage{outlines}
\renewcommand*{\gobble}[1]{}

\usepackage{soul} %
\setstcolor{blue}

\usepackage[dvipsnames]{xcolor}
\usepackage{pgf,tikz,pgfplots}
\PassOptionsToPackage{dvipsnames,table}{xcolor}
\pgfplotsset{compat=1.15}
\usepackage[mathscr]{euscript}
 \let\mathscr\relax%
\usepackage{mathrsfs}
\usetikzlibrary{arrows,backgrounds,shapes}
\usepackage[style=ieee,backend=biber,dashed=false,doi=false,eprint=false, sorting=nyt]{biblatex}
\makeatletter
\def\abx@missing@entry#1{\abx@missing{??}}
\makeatother
\bibliography{main}
\usepackage{pdfpages}
\usepackage{algorithmicx}
\usepackage{algorithm}
\usepackage[noend]{algpseudocode}
\usepackage{amsmath} 
\usepackage{amssymb} 
\usepackage{array}
\usepackage{epsfig}
\usepackage{enumerate}
\usepackage{graphicx} 
\usepackage{pifont}
\usepackage{subcaption}
\usepackage{upgreek}
\usepackage[english]{babel}
\usepackage{csquotes}   %
\usepackage[normalem]{ulem}
\usepackage{xspace}
\usepackage{rotating}
\usepackage{tabularx}
\usepackage{pseudocode}
\usepackage{svg}
\usepackage{tikz}
\usepackage{adjustbox}

\usepackage[T3,T1]{fontenc}
\DeclareSymbolFont{tipa}{T3}{cmr}{m}{n}
\DeclareMathAccent{\invbreve}{\mathalpha}{tipa}{16}

\algrenewcommand\algorithmicrequire{\textbf{Precondition:}}
\algrenewcommand\algorithmicensure{\textbf{Postcondition:}}
\algnewcommand\algorithmicinput{\textbf{Input:}}
\algnewcommand\Input{\item[\algorithmicinput]}
\algnewcommand\algorithmicoutput{\textbf{Output:}}
\algnewcommand\Output{\item[\algorithmicoutput]}
\newtheorem{theorem}{Theorem}[section]
\newenvironment{definition}[1][Definition]{\begin{trivlist}
\item[\hskip \labelsep {\bfseries #1}]}{\end{trivlist}}
\newcolumntype{C}{>{\centering\arraybackslash} m{2.2in} }

\renewcommand{\Re}{\mathbb{R}}

\newcommand{\old}[1]{{}}
\makeatletter
\newcommand{\Rmnum}[1]{\expandafter\@slowromancap\romannumeral #1@}
\makeatother

\newcommand{\jpc}{JPC\xspace}
\newcommand{\jpcs}{JPCs\xspace}

\newcommand\mydots{\ifmmode\ldots\else\makebox[1em][c]{.\hfil.\hfil.}\thinspace\fi}

\usepackage{support-caption}   %
\usepackage[font=footnotesize]{caption}%
\makeatletter
\def\thm@space@setup{%
  \thm@preskip=.5pt plus .5pt minus .5pt
  \thm@postskip=\thm@preskip %
}
\makeatother

\title{\LARGE \bf
Robust-by-Design Plans for Multi-Robot Pursuit-Evasion
}

\author{Trevor Olsen, Nicholas M. Stiffler, and Jason M. O'Kane%
\thanks{%
	T. Olsen and J. M. O'Kane are with the Department
    of Computer Science and Engineering, University of South Carolina, Columbia, SC 29208, USA. N. M. Stiffler is with the Department of Computer Science, University of Dayton, Dayton, OH 45469, USA.
    {\tt \footnotesize tvolsen@email.sc.edu,  jokane@cse.sc.edu, nstiffler1@udayton.edu} %
	This material is based upon work supported by the National Science Foundation 
    under Grant Nos. 1659514 and 1849291.%
}}

\begin{document}

\maketitle
\thispagestyle{empty}
\pagestyle{empty}

\begin{abstract} \pagebudget{0.25}
This paper studies a multi-robot visibility-based pursuit-evasion problem in which a group of pursuer 
robots are tasked with detecting an evader within a two dimensional polygonal environment.
The primary contribution is a novel formulation of the pursuit-evasion problem that modifies
the pursuers' objective by requiring
that the evader still be detected, even in spite of the failure of any single pursuer robot.
This novel constraint, whereby two pursuers are required to detect an evader, 
has the benefit of providing redundancy to the search, should any member of the
team become unresponsive, suffer temporary sensor disruption/failure,  or otherwise become incapacitated.
Existing methods, even those that are designed to %
respond to failures, rely on the 
pursuers to replan and update their search pattern to handle such occurrences. In contrast, the proposed
formulation produces plans that are inherently tolerant of some level of disturbance.
Building upon this new formulation, we introduce an augmented data structure
for encoding the problem state and a novel sampling technique to ensure that the generated plans are robust to failures of any single pursuer robot.
An implementation and simulation results illustrating the effectiveness of this approach are described.

\end{abstract}

\section{Introduction} \pagebudget{0.75}

Pursuit-evasion is a two-player game where the players are diametrically opposed. Members of one team, called the \textit{pursuers}, actively seek out members of the second team, the \textit{evaders}. The pursuers' goal is to capture (either physically or visually) the evaders; the evaders wish to evade capture for as long as possible. 

A number of tasks can be modeled as pursuit-evasion problems albeit with differing levels of antagonism displayed by the evaders. Some such examples include search-and-rescue scenarios (rescuers/rescuee), agriculture/livestock monitoring (stockman/predator), and active pursuit (hunter/prey), among many others. With the proliferation of robots into various domains, we have seen instances where robots are used in all of the aforementioned scenarios to perform the role of the pursuers~\cite{ArnYamTan18, GonDeS16, DiaPar+95}.

This paper addresses a form of the pursuit-evasion problem where the pursuers' task is to establish visibility with all evaders that exist in a given environment. These games take place in a two-dimensional space in which each pursuer is capable of detecting any evader within its line of sight.

\begin{figure}
    \begin{subfigure}[t]{\linewidth}
        \centering
        \includegraphics[width=.5\linewidth]{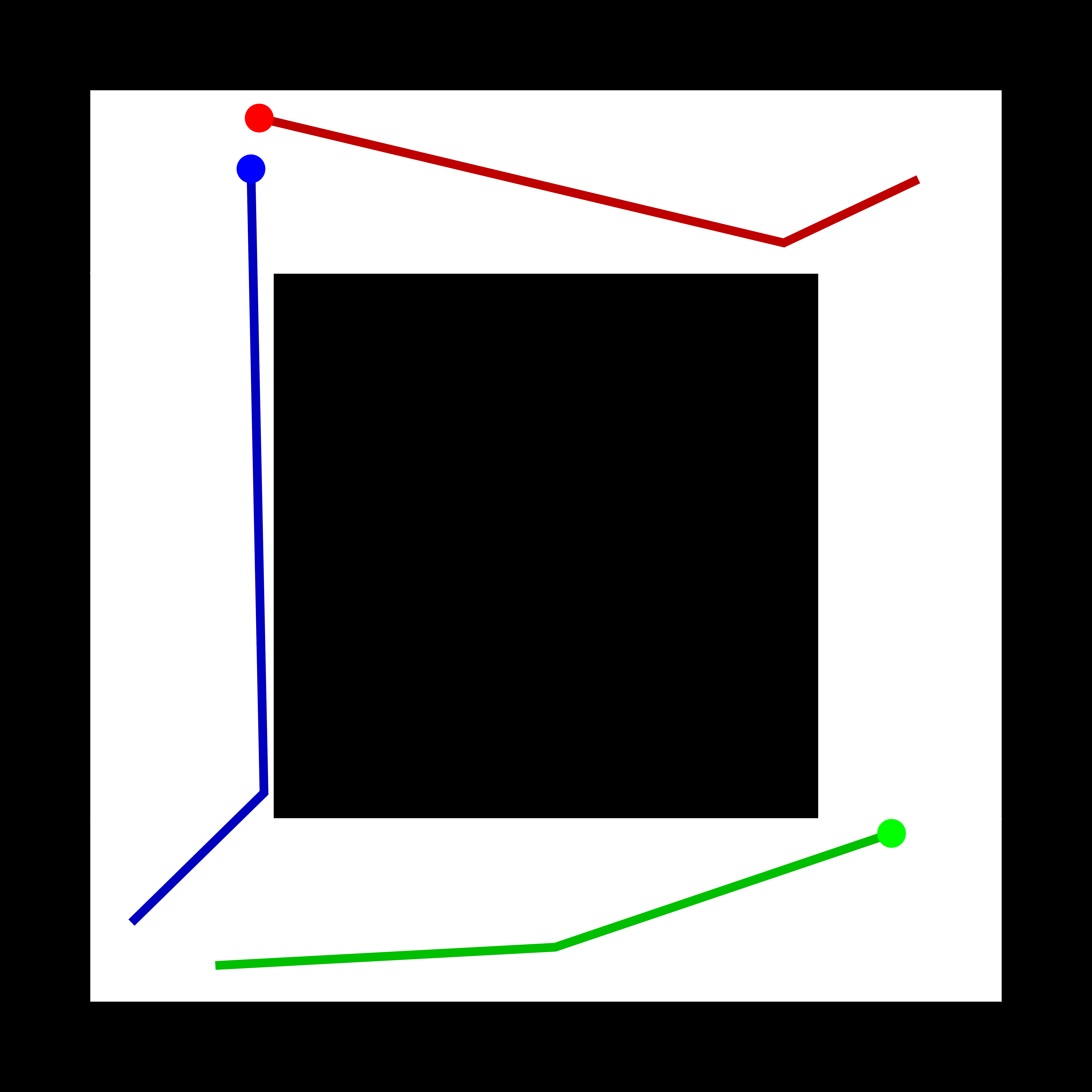}
        \caption{A 1-failure robust solution with 3 pursuers.}
    \end{subfigure}
  \begin{subfigure}[t]{.33\linewidth}
    \centering
    \includegraphics[width=.95\linewidth]{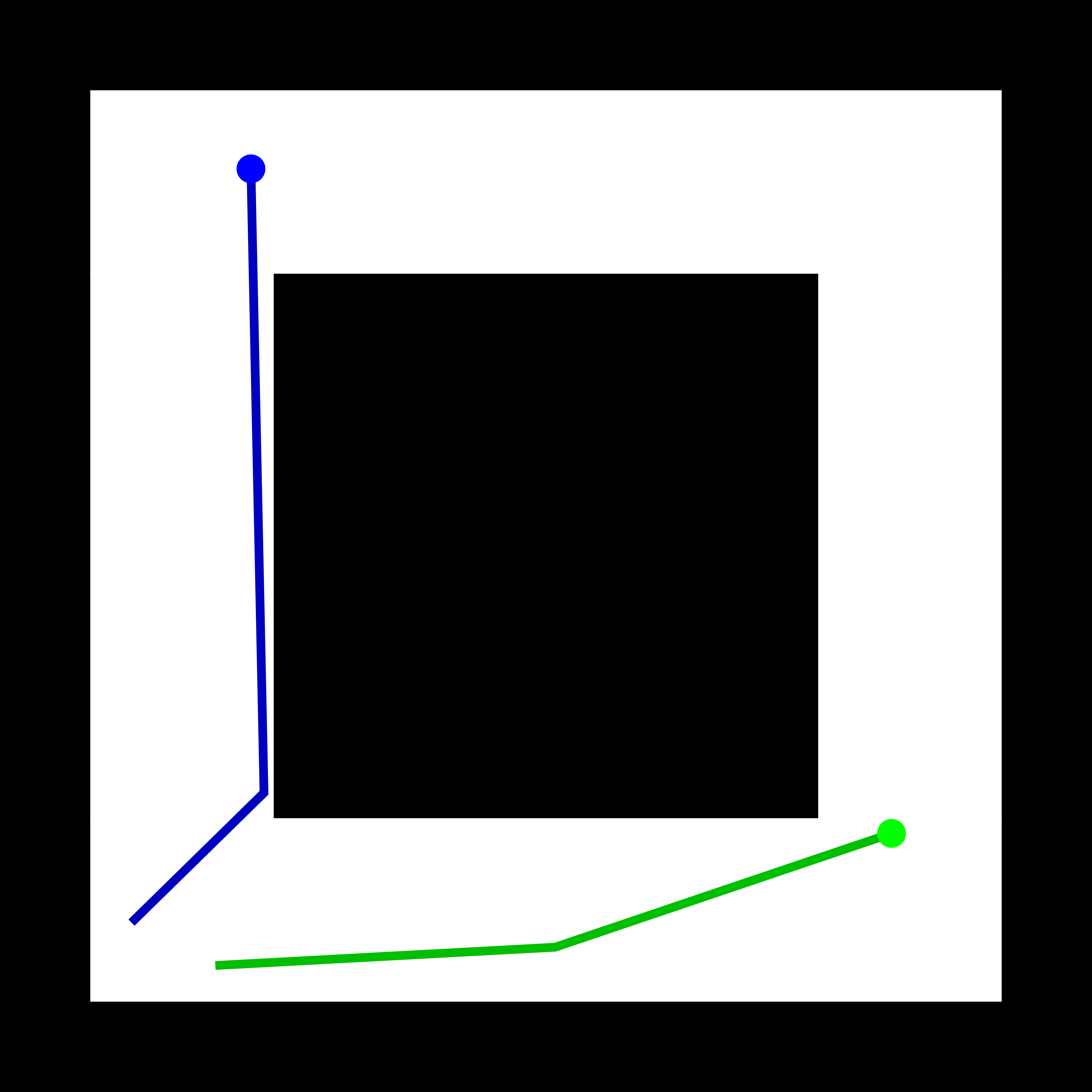}
    \caption{Red failure}
  \end{subfigure}%
  \begin{subfigure}[t]{.33\linewidth}
    \centering
    \includegraphics[width=.95\linewidth]{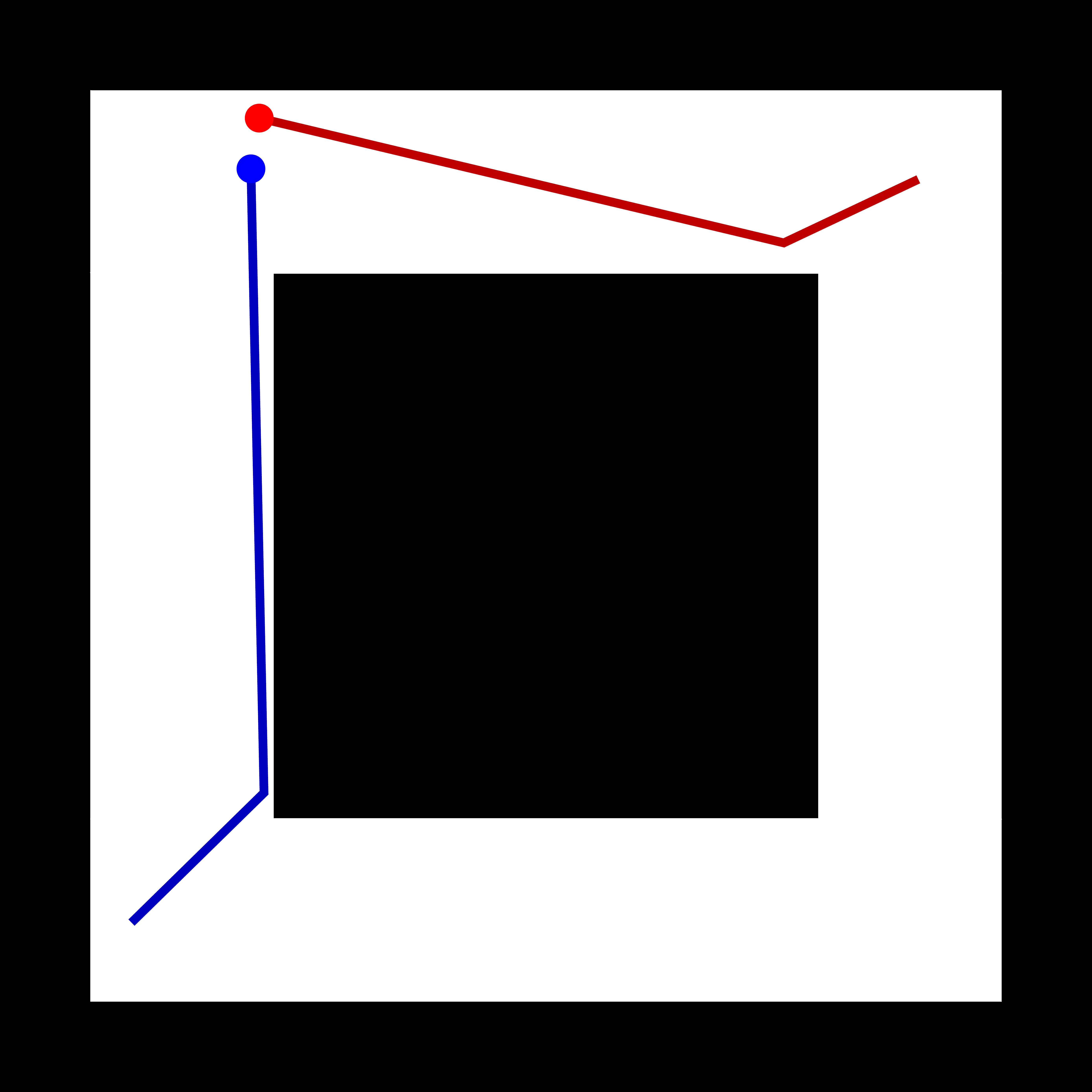}
    \caption{Green failure}
  \end{subfigure}%
  \begin{subfigure}[t]{.33\linewidth}
    \centering
    \includegraphics[width=.95\linewidth]{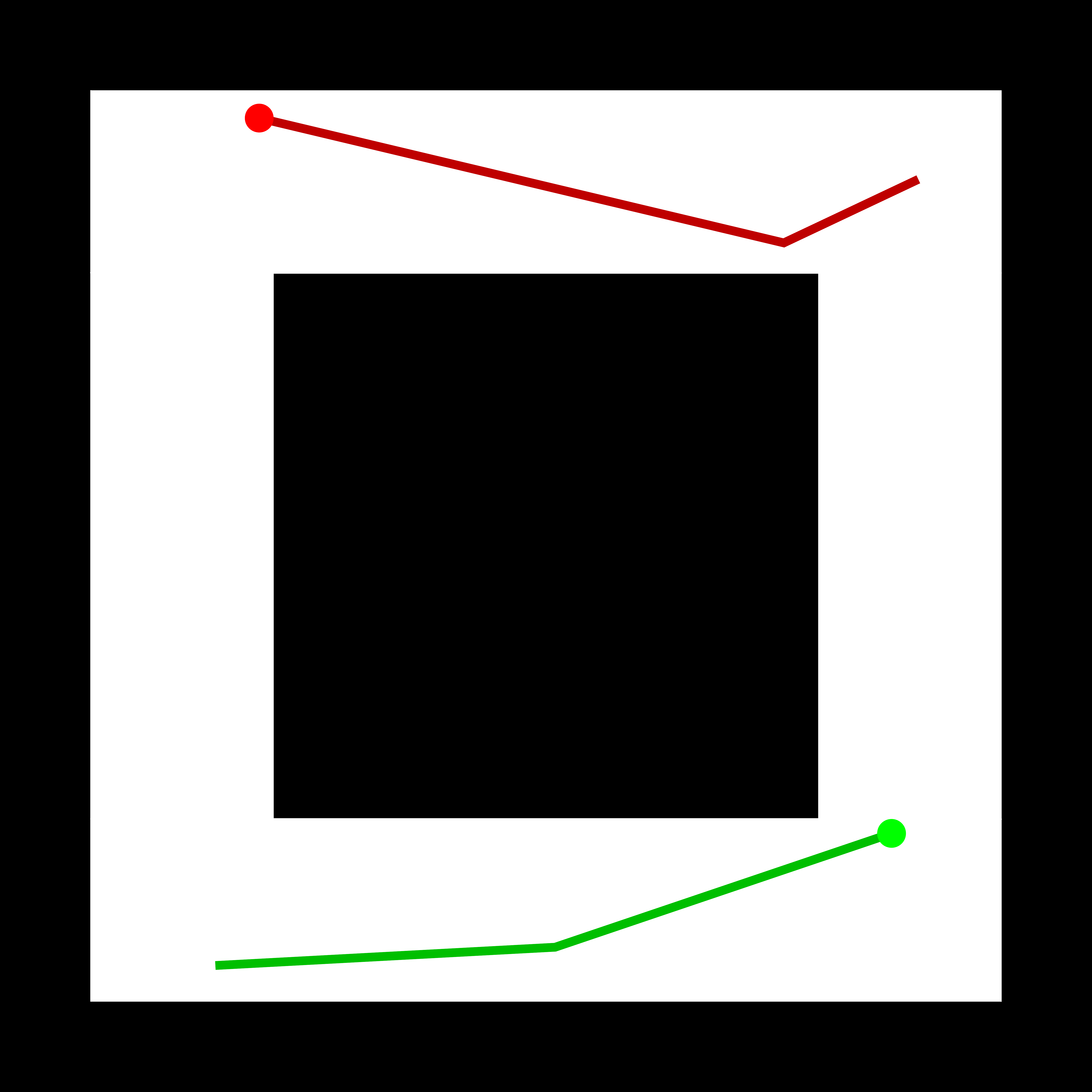}
    \caption{Blue failure}
  \end{subfigure}
  \vspace{-2mm}
  \caption{A simple example of a 1-failure robust solution for $n=3$ robots. Removal of any single pursuer does not preclude the remaining pursuers from capturing an evader in the $n=2$ subsolution.}
  \label{fig:front-page}
\end{figure}

Specifically, this paper investigates how one might generate strategies for the pursuers that are robust in the face of malfunctions which may inhibit members of the team, as shown in Figure~\ref{fig:front-page}. A few such defects include utter failure, intermittent sensor failure (i.e. false negatives), and incorrectly transmitted information. To the authors' knowledge, the literature regarding failures in the realm of pursuit-evasion is scant. Our own previous work considered the situation where members of the pursuer team suffered complete and catastrophic failure, necessitating that the pursuer strategy be re-planned online~\cite{OlsOKaSti21}. This replanning step introduces execution delays and requires all of the remaining pursuers to be aware of failures when and where they occur.  This paper addresses a larger range of potential robotic defects by generating joint motion strategies for the pursuers which guarantee detection of the evaders, regardless of any single pursuer malfunction and without the need to detect such malfunctions when they occur nor to halt the current execution to update the pursuers' motion strategies.

Beyond the previously-studied pursuit-evasion component, which must reason about the regions of the environment which may contain an evader,  the major challenge in generating such `failure-robust' joint pursuer strategies is the need to track these `contaminated' regions separately for each possible pursuer failure, while ensuring that a single solution can be extracted when the process is complete.
Note that the best known complete algorithm  for multi-robot visibility-based pursuit-evasion solves the problem in  time doubly exponential in the number of pursuer agents. To address this shortcoming, sampling-based methods have been developed~\cite{StiOKa14b,OlsTum+21} that utilize a graph structure whose vertices encode the pursuers pose information in addition to the regions of the environment where the evader may be; whereas edges indicate feasible pursuer motions within the environment.

This paper builds on this line of sampling-based approaches by providing an augmented graph structure that encodes the "robustness" of the search (i.e. reasoning over any possible pursuer failure). In response to this added layer of complexity, a novel sampling method is introduced which creates condensed sample sets that ensure capture of the evader via a highly connected roadmap.

The remainder of the article begins with a review the related literature in Section~\ref{sec:related}. Next, a formal description of the problem is provided in Section~\ref{sec:ps}. Section~\ref{sec:ad} outlines the algorithm utilized to generate robust joint motion strategies. Simulation results, in which the algorithm was employed in several different representative environments, are described in Section~\ref{sec:experiments} before the paper concludes with a summary and look towards future work Section~\ref{sec:conclusion}.

\section{Related work}\label{sec:related} \pagebudget{0.75}

Pursuit-evasion problems have a rich history in the fields of differential game theory~\cite{HoBryBar65, Isa65}, graph theory~\cite{Par76}, and geometric settings~\cite{SuzYam92,GuiLat+99}. In the context of graph theory, these problems are typically referred to as discrete pursuit-evasion games. One possible scenario for such a game requires the pursuer agent(s) to occupy the same vertex as the evader. Parsons introduced this problem on finite graphs \cite{Par76}. Interest in graph theoretic pursuit-evasion problems grew quickly, spanning a broad range of variations such as: unrestricted evader movements \cite{Adl+02} and infinite graphs \cite{Leh16}. 

Pursuit-evasion also has a rich history in the geometric/continuous domain \cite{SkhKakDud18, ZouBha19, IslSunSas05}. The continuous nature of geometric environments
allows for several different types of capture conditions, including those based on visibility~\cite{SuzYam92}. The visibility-based pursuit-evasion game has been studied in extensive detail for both the case of a single pursuer as well as a team of several pursuers. For the single pursuer case, results include: completeness \cite{GuiLat+99}, solvability conditions \cite{ParLeeChwa01}, limited field of view \cite{GerThrGor06}, bounded evader velocity \cite{TovLav08}, optimality \cite{StiOKa17} and robustness \cite{StiOKa20}. 

Contrary to what one might expect, the addition of extra pursuer agents does not necessarily make the problem any easier. As the number of pursuers increases, the dimensionality of the problem increases creating a computational challenge.
This concern with dimensionality has parallels in many areas of robotics.
In path and motion planning, for example, sampling-based approaches have proven remarkably useful~\cite{KarFra11, KinMolKav18}.
This approach has also emerged in multi-robot pursuit-evasion problems, where sampling \cite{StiOKa14a, OlsTum+21} was able to successfully combat the complexity of a complete algorithm \cite{StiOKa14b}. Other research in multi-robot pursuit-evasion includes the lack of environmental knowledge \cite{KolCar10a} and heuristics for the worst-case \cite{GreGiv+17}.

Fault tolerance and recovery is an active research thread in the robotics community as well \cite{Cre+15, Vin+94, Don88}, but less so in the context of pursuit-evasion. The authors previously introduced a scheme to handle complete and catastrophic failures~\cite{OlsOKaSti21}.
The novelty of this paper is to address the situation where one of the pursuers is assumed to be faulty, without the need to detect the fault nor replan on the fly. In particular, we consider malfunctions such as complete failure with and without indication, potential erroneous information reporting, whether it be malicious or accidental and possible physical limitations leaving the robot immobilized for the remainder of the search.

\section{Problem statement}\label{sec:ps} \pagebudget{1.25}

\subsection{The Environment, Pursuers and Evaders}

The environment, $F$, is a closed, bounded and polygonal subset of $\Re^2$. 
A team of $n$ pursuers move continuously in $F$ at a bounded speed. We denote the location of the $i^{\text{th}}$ pursuer
at time $t$ by the continuous function $f_i(t): [0, \infty) \rightarrow F$. We call each such function $f_i$ a motion strategy. Given $n$ pursuers and their motion strategies, we call the vector $\langle f_1(t), \dots , f_n(t) \rangle$ the joint pursuer configuration (\jpc) at time $t$. Each pursuer is equipped with
an omnidirectional sensor which extends to the nearest point on the boundary of the environment.
For a pursuer located at $q \in F$, the visibility polygon, $V(q)$, is the set of points $r \in F$ such that the line segment $\overline{qr}$ is contained in $F$.

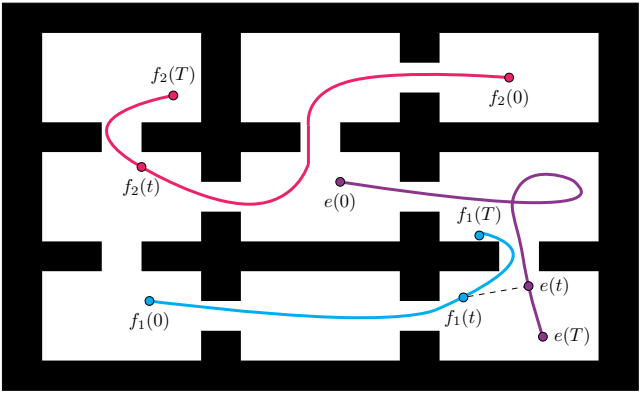
\begin{figure}[t]
    \centering
    
    \resizebox{0.99\columnwidth}{!}{
    \begin{tikzpicture}[scale=0.75, yscale=0.75]
        \fill[draw=black,fill=black] (-1,-1) rectangle (15,12);
        \fill[color=white, line width=0.3mm] (0,0)--(0,3)--(1.5,3)--(1.5,4)--(0,4)--(0,7)--(1.5,7)--(1.5,8)--(0,8)--(0,11)--(4,11)--(4,8)--(2.5,8)--(2.5,7)--(4,7)--(4,6)--(5,6)--(5,7)--(6.5,7)--(6.5,8)--(5,8)--(5,11)--(9,11)--(9,10)--(10,10)--(10,11)--(14,11)--(14,8)--(10,8)--(10,9)--(9,9)--(9,8)--(7.5,8)--(7.5,7)--(9,7)--(9,6)--(10,6)--(10,7)--(14,7)--(14,4)--(12.5,4)--(12.5,3)--(14,3)--(14,0)--(10,0)--(10,1)--(9,1)--(9,0)--(5,0)--(5,1)--(4,1)--(4,0);
        \fill[color=black] (11.5,4)--(10,4)--(10,5)--(9,5)--(9,4)--(5,4)--(5,5)--(4,5)--(4,4)--(2.5,4)--(2.5,3)--(4,3)--(4,2)--(5,2)--(5,3)--(9,3)--(9,2)--(10,2)--(10,3)--(11.5,3);

        \draw [WildStrawberry, ultra thick] plot [smooth, tension=0.9] coordinates {(11.75, 9.5) (7.8,9.4) (6.7, 7.9)};
        \draw [WildStrawberry, ultra thick] plot [smooth, tension=1] coordinates {(6.7, 7.9) (6.7, 6.6)};
        \draw [WildStrawberry, ultra thick] plot [smooth, tension=1] coordinates {(6.7,6.6) (5.25,5.25) (2.5,6.5)};
        \draw [WildStrawberry, ultra thick] plot [smooth, tension=1] coordinates {(2.5,6.5) (1.65,7.9) (3.3,8.9)};
        
        \node[fill=WildStrawberry, circle, draw=black, scale=0.5, label=below:{$f_2(0)$}] (q1) at (11.75,9.5) {};
        \node[fill=WildStrawberry, circle, draw=black, scale=0.5, label=below:{$f_2(t)$}] (qt) at (2.5,6.5) {};
        \node[fill=WildStrawberry, circle, draw=black, scale=0.5, label=above:{$f_2(T)$}] (q2) at (3.3,8.9) {};

        \draw [cyan, ultra thick] plot [smooth, tension=1] coordinates {(2.7,2) (7.4,1.45) (9.9,1.7)};
        \draw [cyan, ultra thick] plot [smooth, tension=1] coordinates {(9.9,1.7) (11.8,3.3) (11,4.3)};
        
        \node[fill=cyan, circle, draw=black, scale=0.5, label=below:{$f_1(0)$}] (p1) at (2.7,2) {};
        \node[fill=cyan, circle, draw=black, scale=0.5, label=below:{$f_1(t)$}] (pt) at (10.6,2.12) {};
        \node[fill=cyan, circle, draw=black, scale=0.5, label=above:{$f_1(T)$}] (p2) at (11,4.2) {};

        \draw [Fuchsia, ultra thick] plot [smooth, tension=1] coordinates {(7.5,6) (12.4,5.3) (13.4,6) (12.0,5.85) (12.1,3.45) (12.6,0.8)};
        
        \node[fill=Fuchsia, circle, draw=black, scale=0.5, label=below:{$e(0)$}] (e1) at (7.5,6) {};        
        \node[fill=Fuchsia, circle, draw=black, scale=0.5, label=right:{$e(t)$}] (et) at (12.24,2.5) {};
        \node[fill=Fuchsia, circle, draw=black, scale=0.5, label=right:{$e(T)$}] (ett) at (12.6,0.8) {};
        \draw [color=black,dashed] (pt)--(et);
    \end{tikzpicture}}
    \caption{Two pursuers actively search for the evader for $T$ seconds. At time $t$, the pursuer moving along path $f_1$ has visibility of the evader.} 
    \label{fig:notation}
\end{figure}

Similar to pursuers, evaders move continuously throughout $F$ but differ in that they are capable of reaching arbitrarily
high speeds. Since the objective of the pursuers is to locate the evaders regardless of the trajectories taken by the 
evaders, we can assume, without loss of generality, there exists a single evader.
We denote the location of this evader at time $t$ by the
continuous function $e(t): [0, \infty) \rightarrow F$.
The pursuers' objective is to guarantee to locate the evader, as formalized in the next definition.
\begin{definition}
A collection of motion strategies $X = \langle f_1, \dots , f_n \rangle$ is called a \emph{solution} if, for any evader curve $e$, there exists a time $t \geq 0$ such that $e(t) \in \bigcup_{i \leq n} V(f_i(t))$.
\end{definition}
Figure~\ref{fig:notation} exemplifies these definitions.
We are interested in solutions that can still guarantee the detection of the evader, even if a single pursuer robot fails.

\begin{definition} Given a set of robots, $R$, where $|R| > k$, a solution $X$ is \emph{$k$-failure robust} if it remains a solution utilizing only the robots $R \setminus B$, for any $B \subset R$ with $|B| \leq k$.  
\end{definition}
Notice that, according to these definitions, a solution is a 0-failure robust solution. In this paper, we address the problem of generating 1-failure robust solutions.
 
\subsection{Shadows and Their Events}
The ideas presented in the following two subsections, which discuss the unseen regions of the environment, were initially presented by Guibas, Latombe, LaValle, Lin, and Motwani~\cite{GuiLat+99} in the context of a single-robot version of this problem. Because our algorithm, specifically the rSG-PEG data structure described in Section~\ref{sec:rSG-PEG}, uses these ideas in important ways, we summarize them here; details appear in the original paper.

\subsubsection{Shadows}
For a fixed time $t$, the \emph{shadow region} is the set $S(t) = F \setminus \bigcup_{i \leq n} V(f_i(t))$. The maximally connected components of the shadow region are called \emph{shadows}. Based on the pursuers' movement up until time $t$, it is either possible or not for an evader to be in a certain shadow while still undetected. In the case that a shadow could contain an undetected evader, the \emph{status} of the shadow is \emph{cleared}, otherwise the shadow's status is \emph{contaminated}. We concisely summarize the status of each shadow at a given time by the \emph{shadow label}, a binary string of length equal to the number of shadows. In a shadow label, a 1 bit represents a contaminated shadow, while a 0 bit indicates that the corresponding shadow is cleared. A shadow label consisting of all zeroes indicates that all shadows have been cleared, making it impossible for an undetected evader to be anywhere within the environment, thus resulting in a solution.

It will be useful to consider a notion of dominance between shadow labels at the same JPC.  For two shadow labels $\ell = (a_1 a_2 \cdots a_m)$ and $\ell' = (b_1 b_2 \cdots b_m)$, $\ell$ is said to \emph{dominate} $\ell'$ if $\ell \ne \ell'$ and for each $i = 1, \dots, m$, $a_i \leq b_i$. That is, if a shadow is contaminated in $\ell$, it must also be contaminated in $\ell'$ and any cleared shadow in $\ell'$ must also be cleared in $\ell$.

 \subsubsection{Shadow Events}\label{sec:shadowevents}
 As pursuers move within $F$, the shadows will continually change shape and size. When the number of shadows changes, these occurrences are called \emph{shadow events}, which can be classified into four types:
 
 \begin{itemize}
     \item \emph{Appear}: A shadow can appear if a previously seen subset of the environment falls out of the visibility polygon of all of the pursuers. In this case, the newly formed shadow is marked as clear.
     \item \emph{Disappear}: If a pursuer gains vision of an entire shadow, the shadow disappears. Here, the shadow is completely removed from the shadow label.
     \item \emph{Merge}: Two or more shadows merge if they become a single connected component. When this occurs, the newly merged shadow is given the cleared label if and only if all of the merging are cleared. 
     \item \emph{Split}: If a pursuer's visibility polygon disconnects an existing shadow, we say the shadow was split. Both post-split shadows are given the status of the pre-split shadow.
 \end{itemize}

\section{Algorithm Overview}\label{sec:ad} \pagebudget{1.75}
This section describes an algorithm to generate a 1-failure robust solution for a given environment and number of pursuers.

The algorithm builds upon earlier sampling-based approaches to visibility-based pursuit-evasion~\cite{StiOKa14b}.
The basic idea in that prior work is to construct a roadmap data structure
that represents the pursuers' ability to move through the environment and to clear various collections of shadows.
As \jpcs are sampled and inserted into the roadmap, the prior algorithm tracks a set of reachable shadow labels attached to each vertex.  When the data structure determines that an all-clear shadow label is reachable at some vertex, the algorithm extracts that solution and terminates successfully.

The algorithm we propose here differs from that baseline in three important ways, necessitated by the need to produce robust solutions.
First, the data structure is augmented to track shadow labels not for all $n$ robots, but instead for each of the $n$ distinct subsets of size $n-1$.  This ensures that, if \emph{all} of these shadow labels achieve an all clear status, the resulting solution will be $1$-failure robust.  See Section~\ref{sec:rSG-PEG}.  Second, we introduce a geometric caching optimization to the data structure, based on the observation that the number of times shadow labels are propagated across each edge is substantially higher than in prior settings. Section~\ref{sec:cache} describes this change.  Finally, we introduce a new sampling scheme tailored to the specific need for solutions in which at least two robots clear each shadow (Section~\ref{sec:sampling}). 

\subsection{Robust Sample-Generated Pursuit-Evasion Graphs (rSG-PEG)}\label{sec:rSG-PEG}
Here, we describe how we enhanced the existing sample-generated pursuit-evasion graph (SG-PEG) data structure~\cite{StiOKa14b} to generate 1-failure robust solutions.
An rSG-PEG, $G$, is a directed graph in which one vertex is designated as the root.  Each vertex of $G$ is labeled with a \jpc; a directed edge $u \to w$ indicates 
that there exists a coordinate-wise straight line, collision free movement between the \jpcs of $u$ and $w$. For the sake of compactness, a vertex with \jpc $w_1, \dots, w_n$ will be named $w$.

Each vertex is also associated with a collection of \emph{failure shadow labels}, each of which is an $n$-tuple $L=(\ell_1, \ldots, \ell_n)$, where each $\ell_i$ is a single shadow label.  The interpretation is that the existence of failure shadow label $(\ell_1, \ldots, \ell_n)$ at vertex $w$ implies that 
there exists a path in $G$ from the root to $w$, such that for each $1 \le i \le n$, the pursuers in $\{1, \ldots n \} \setminus \{ i \}$ would reach shadow label $\ell_i$ by following that path.
Thus, reaching a vertex with a failure shadow label in which all $n$ sub-labels are all cleared results in a 1-failure robust solution within $G$.

An rSG-PEG is constructed by repeated calls to its primary operation, $G.\textsc{AddSample}(w)$, which performs the following steps.
\begin{enumerate}
    \item A vertex $w$ is added to $G$.
    \item An edge $w \to u$ is added between between $w$ and each other vertex $u$ of $G$ if the line segment $\overline{wu}$ is fully contained in $F^n$. Similarly the twin edge $u \to w$ is added to $G$.
    \item During this process, each time an edge $a \to b$ is created, for each failure shadow label $L$ attached to $a$, the algorithm computes a failure shadow label $L'$ for $b$, computed by propagating each shadow label $\ell_i$ in $L$ across the edge $a \to b$, as described below.  If $L'$ is not dominated by any other failure shadow label at $b$, it is retained at $b$.  Here, dominance of failure shadow labels is defined by generalizing the idea of dominance of shadow labels.  This process continues recursively, spreading new reachable failure shadow labels across $G$ as needed.
\end{enumerate}
Adding a sequence of samples to an rSG-PEG is enough to form a 1-failure robust solution.  However,
propagating the shadow information along edges of the rSG-PEG is a computationally expensive operation that occurs frequently in the robust-failure scenarios considered here.  The next section describes how to eliminate redundant geometric computations from this process.

\subsection{Fast Label Propagation via Shadow Influence Caching}\label{sec:cache}
To accelerate the propagation of shadow labels across edges of the rSG-PEG, we construct 
shadow influence relations for each edge.  For a given edge $w \to u$ and a given pursuer index $i$, let $S_{w,i}$ and $S_{u,i}$ denote the shadows at $w$ and $u$ formed in the absence of pursuer $i$. The \emph{shadow influence relation} is a relation $R \subseteq S_{w,i} \times S_{u,i}$, in which $(s_w, s_u) \in R$ if and only if there exists a trajectory for the evader to travel undetected from $s_w$ to $s_u$ as the pursuers move from $w$ to $u$.
These shadow influence relations can be computed according to the update rules described in Section~\ref{sec:shadowevents}.  This is a time-consuming computational geometry operation, but it must be performed only once for each edge-pursuer pair.
Once the relation $R$ is computed, any future shadow label can be propagated efficiently by setting each shadow $s_u$ at $u$ to be contaminated if and only if there exists a contaminated shadow $s_w$ in the source shadow label for which $(s_w, s_u) \in R$.

\begin{figure}

  \begin{subfigure}[t]{.49\columnwidth}
    \centering
    \includegraphics[width=1\columnwidth]{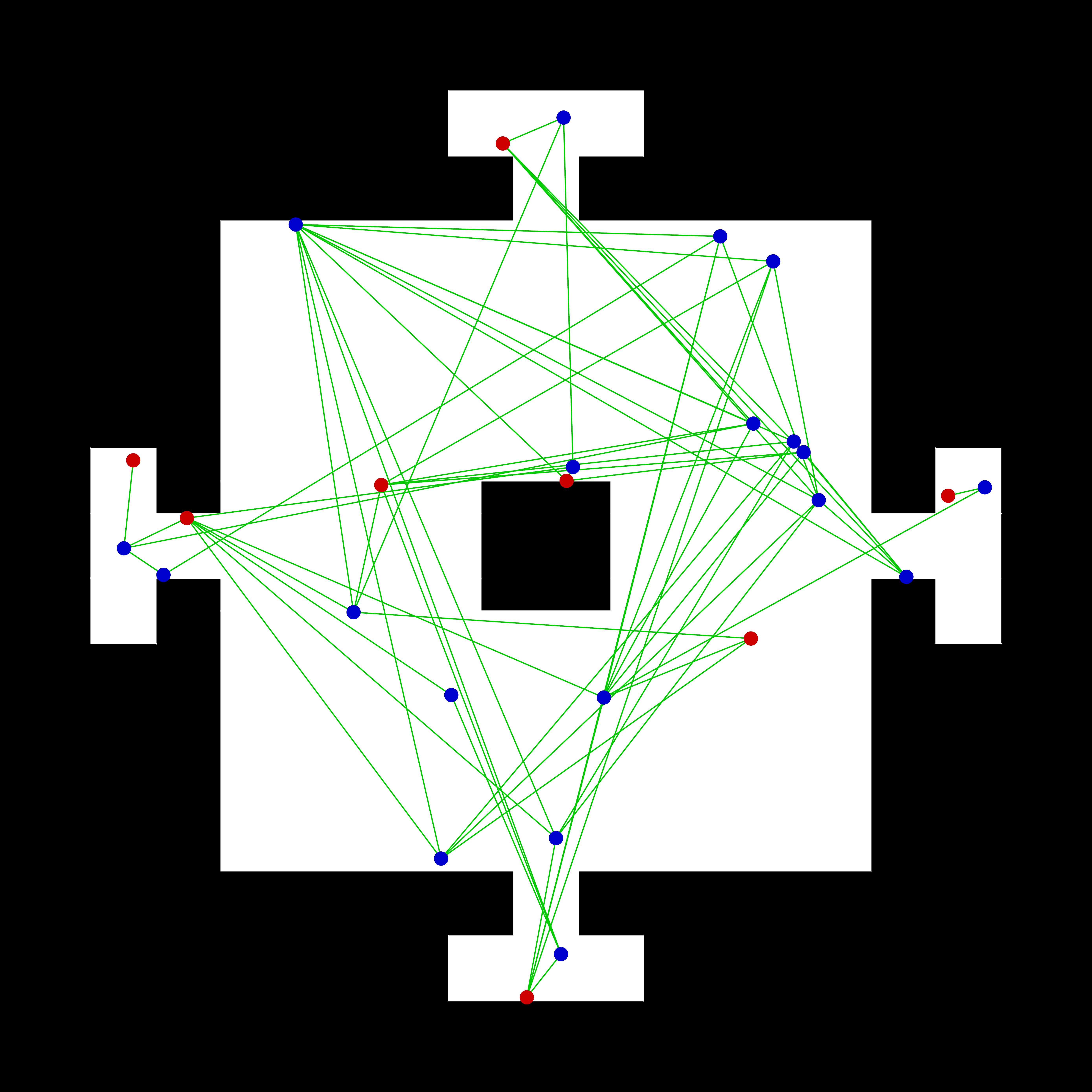}
    \caption{Original web.}
  \end{subfigure}
  \begin{subfigure}[t]{.49\columnwidth}
    \centering
    \includegraphics[width=1\columnwidth]{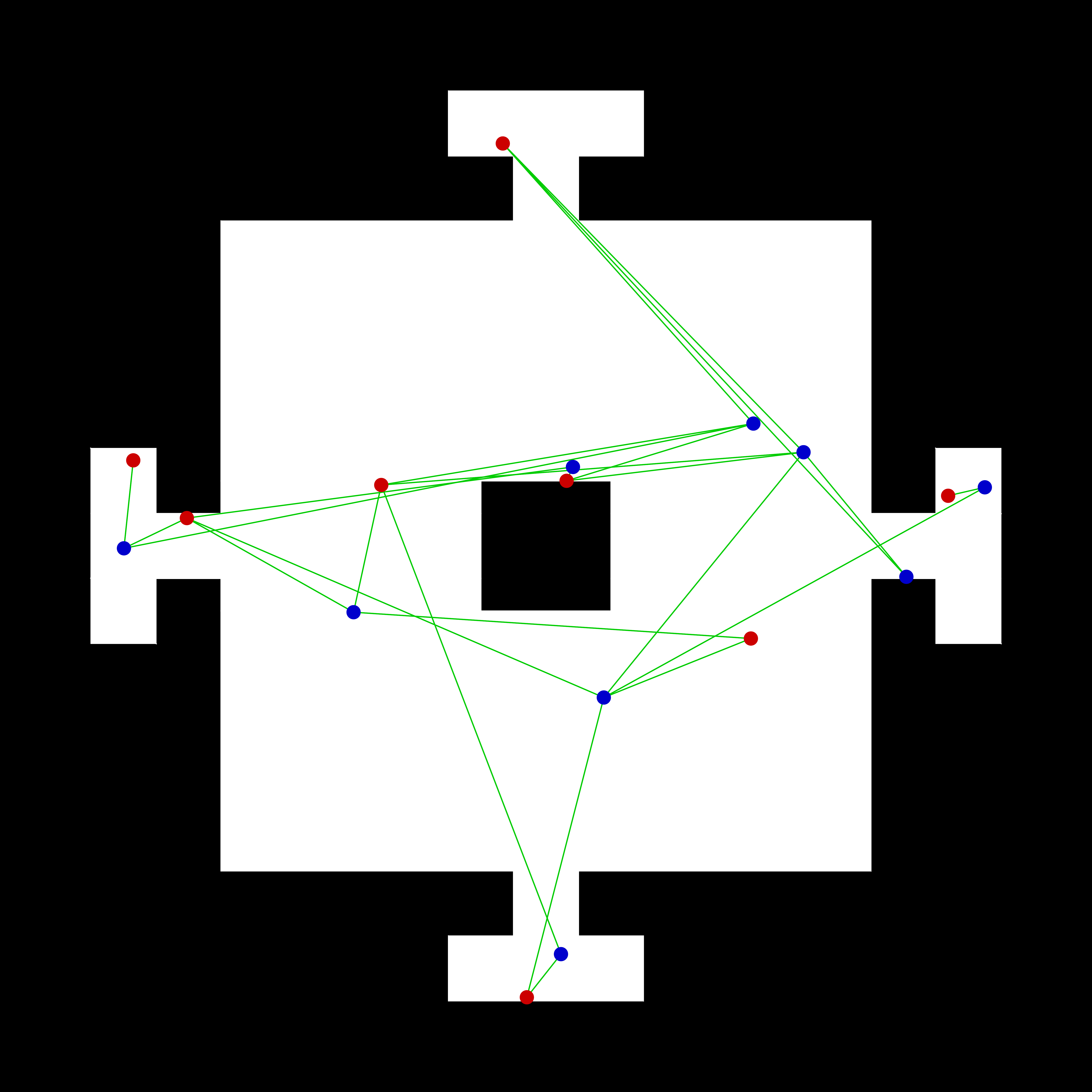}
    \caption{Sparse web.}
  \end{subfigure}
  \caption{A visualization of the noise reduction in sparse webs. The red points represent the initial points while the intersection points are drawn in blue}
  \label{fig:webs}
\end{figure}

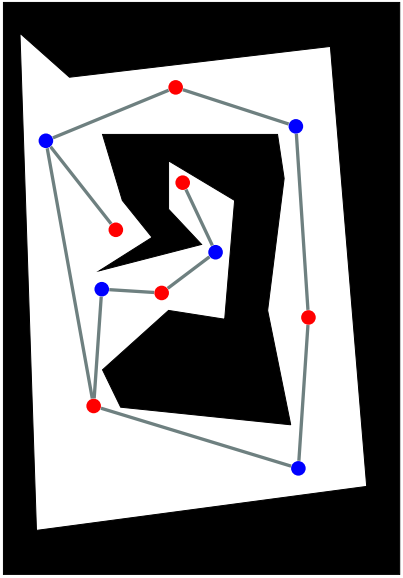
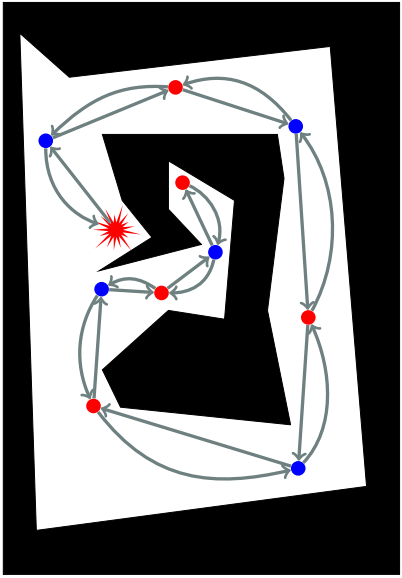
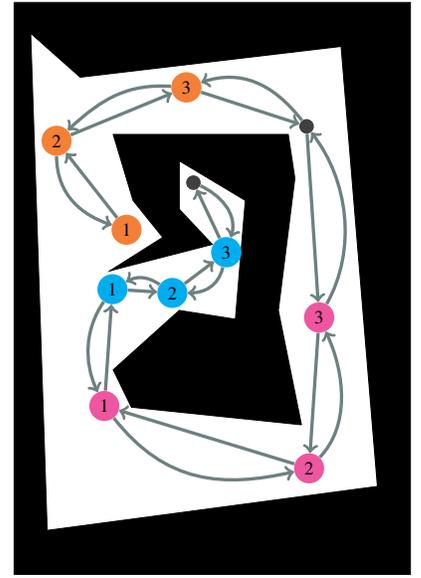
\begin{figure*}[t]

  \begin{subfigure}[t]{.31\textwidth}
    \centering
    \begin{adjustbox}{rotate=90}
        \centering
    \resizebox{1.4\textwidth}{!}{
    \begin{tikzpicture}[scale=0.75, yscale=0.75]
        \fill[draw=black,fill=black] (-2,-1) rectangle (11,11); 
        \fill[draw=black,fill=white] (0,0)--(10,1.1)--(9.3,9)--(10.3,10.5)--(-1,10)--(0,0);
        \fill[draw=black,fill=black] (1.4,2.3)--(4,3)--(7,2.5)--(8,2.7)--(8,8)--(6.5,7.4)--(5.66, 6.5)--(4.9,8.1)--(5.5,5)--(6.3,6)--(7.4,6)--(6.5,4)--(3.8,4.3)--(4,6)--(2.65,8)--(1.8,7.45)--(1.4,2.3);

        \node[fill=red, circle, scale=0.75] (p1) at (3.84, 1.764) {};
        \node[fill=red, circle, scale=0.75] (p2) at (9.06954, 5.785) {};
        \node[fill=red, circle, scale=0.75] (p3) at (1.82804, 8.2682) {};
        \node[fill=red, circle, scale=0.75] (p4) at (6.9038, 5.572) {};
        \node[fill=red, circle, scale=0.75] (p5) at (4.39675, 6.20912) {};
        \node[fill=red, circle, scale=0.75] (p6) at (5.8324, 7.5966) {};
        
        \node[fill=blue, circle, scale=0.75] (q1) at (8.1833, 2.14469) {};
        \node[fill=blue, circle, scale=0.75] (q2) at (0.410033, 2.0689) {};
        \node[fill=blue, circle, scale=0.75] (q3) at (7.85585, 9.71485) {};
        \node[fill=blue, circle, scale=0.75] (q4) at (4.48226, 8.02311) {};
        \node[fill=blue, circle, scale=0.75] (q5) at (5.3225, 4.57663) {};
        
        \definecolor{darkgray}{rgb}{0.43, 0.5, 0.5}
        
        \draw [color=darkgray, ultra thick] (p4)--(q5);
        \draw [color=darkgray, ultra thick] (q5)--(p5);
        \draw [color=darkgray, ultra thick] (p5)--(q4);
        \draw [color=darkgray, ultra thick] (q4)--(p3);
        \draw [color=darkgray, ultra thick] (p3)--(q2);
        \draw [color=darkgray, ultra thick] (q2)--(p1);
        \draw [color=darkgray, ultra thick] (p1)--(q1);
        \draw [color=darkgray, ultra thick] (q1)--(p2);
        \draw [color=darkgray, ultra thick] (p2)--(q3);
        \draw [color=darkgray, ultra thick] (q3)--(p6);
        \draw [color=darkgray, ultra thick] (q3)--(p3);
    
    \end{tikzpicture}}
    \end{adjustbox}
    \caption{A sparse web.}
  \end{subfigure}
  \hspace{0.02\textwidth}
  \begin{subfigure}[t]{.31\textwidth}
  \centering
    \begin{adjustbox}{rotate=90}
        \centering
    \resizebox{1.4\columnwidth}{!}{
    \begin{tikzpicture}[scale=0.75, yscale=0.75]
        \fill[draw=black,fill=black] (-2,-1) rectangle (11,11); 
        \fill[draw=black,fill=white] (0,0)--(10,1.1)--(9.3,9)--(10.3,10.5)--(-1,10)--(0,0);
        \fill[draw=black,fill=black] (1.4,2.3)--(4,3)--(7,2.5)--(8,2.7)--(8,8)--(6.5,7.4)--(5.66, 6.5)--(4.9,8.1)--(5.5,5)--(6.3,6)--(7.4,6)--(6.5,4)--(3.8,4.3)--(4,6)--(2.65,8)--(1.8,7.45)--(1.4,2.3);
        
        \node[fill=red, circle, scale=0.75] (p1) at (3.84, 1.764) {};
        \node[fill=red, circle, scale=0.75] (p2) at (9.06954, 5.785) {};
        \node[fill=red, circle, scale=0.75] (p3) at (1.82804, 8.2682) {};
        \node[fill=red, circle, scale=0.75] (p4) at (6.9038, 5.572) {};
        \node[fill=red, circle, scale=0.75] (p5) at (4.39675, 6.20912) {};
        \node[fill=red, starburst, scale=0.75] (p6) at (5.8324, 7.5966) {};
        
        \node[fill=blue, circle, scale=0.75] (q1) at (8.1833, 2.14469) {};
        \node[fill=blue, circle, scale=0.75] (q2) at (0.410033, 2.0689) {};
        \node[fill=blue, circle, scale=0.75] (q3) at (7.85585, 9.71485) {};
        \node[fill=blue, circle, scale=0.75] (q4) at (4.48226, 8.02311) {};
        \node[fill=blue, circle, scale=0.75] (q5) at (5.3225, 4.57663) {};
        
        \definecolor{darkgray}{rgb}{0.43, 0.5, 0.5}
        
        \draw [color=darkgray, ->, style=ultra thick] (p6)--(q3);
        \draw [color=darkgray, ->, style=ultra thick] (q3)--(p2);
        \draw [color=darkgray, ->, style=ultra thick] (p2)--(q1);
        \draw [color=darkgray, ->, style=ultra thick] (q1)--(p1);
        \draw [color=darkgray, ->, style=ultra thick] (p1)--(q2);
        \draw [color=darkgray, ->, style=ultra thick] (q2)--(p3);
        \draw [color=darkgray, ->, style=ultra thick] (p3)--(q4);
        \draw [color=darkgray, ->, style=ultra thick] (q4)--(p5);
        \draw [color=darkgray, ->, style=ultra thick] (p5)--(q5);
        \draw [color=darkgray, ->, style=ultra thick] (q5)--(p4);
        
        \draw [color=darkgray, ->, style=ultra thick] (p4) to [out=250,in=340] (q5);
        \draw [color=darkgray, ->, style=ultra thick] (q5) to [out=165,in=270] (p5);
        \draw [color=darkgray, ->, style=ultra thick] (p5) to [out=60,in=295] (q4);
        \draw [color=darkgray, ->, style=ultra thick] (q4) to [out=140,in=30] (p3);
        \draw [color=darkgray, ->, style=ultra thick] (p3) to [out=225,in=100] (q2);
        \draw [color=darkgray, ->, style=ultra thick] (q2) to [out=310,in=210] (p1);
        \draw [color=darkgray, ->, style=ultra thick] (p1) to [out=325,in=225] (q1);
        \draw [color=darkgray, ->, style=ultra thick] (q1) to [out=45,in=285] (p2);
        \draw [color=darkgray, ->, style=ultra thick] (p2) to [out=87,in=307] (q3);
        \draw [color=darkgray, ->, style=ultra thick] (q3) to [out=180,in=75] (p6);
    
    \end{tikzpicture}}
    \end{adjustbox}
    \caption{To form the walk $D$, DFS is performed on the sparse web starting at the star vertex. In this example, the length the walk $D$ is $d=20$.}
  \end{subfigure}
  \hspace{0.02\textwidth}
\begin{subfigure}[t]{.31\textwidth}
  \centering
    \begin{adjustbox}{rotate=90}
        \centering
    \resizebox{1.4\columnwidth}{!}{
    \begin{tikzpicture}[scale=0.75, yscale=0.75, rotate=0]
        \fill[draw=black,fill=black] (-2,-1) rectangle (11,11); 
        \fill[draw=black,fill=white] (0,0)--(10,1.1)--(9.3,9)--(10.3,10.5)--(-1,10)--(0,0);
        \fill[draw=black,fill=black] (1.4,2.3)--(4,3)--(7,2.5)--(8,2.7)--(8,8)--(6.5,7.4)--(5.66, 6.5)--(4.9,8.1)--(5.5,5)--(6.3,6)--(7.4,6)--(6.5,4)--(3.8,4.3)--(4,6)--(2.65,8)--(1.8,7.45)--(1.4,2.3);

        \node[fill=VioletRed, circle, scale=0.75] (p1) at (3.84, 1.764) {\rotatebox{-90}{\large3}};
        \node[fill=Orange, circle, scale=0.75] (p2) at (9.06954, 5.785) {\rotatebox{-90}{\large3}};
        \node[fill=VioletRed, circle, scale=0.75] (p3) at (1.82804, 8.2682) {\rotatebox{-90}{\large1}};
        \node[fill=darkgray, circle, scale=0.75] (p4) at (6.9038, 5.572) {};
        \node[fill=Cyan, circle, scale=0.75] (p5) at (4.39675, 6.20912) {\rotatebox{-90}{\large2}};
        \node[fill=Orange, circle, scale=0.75] (p6) at (5.8324, 7.5966) {\rotatebox{-90}{\large1}};
        
        \node[fill=darkgray, circle, scale=0.75] (q1) at (8.1833, 2.14469) {};
        \node[fill=VioletRed, circle, scale=0.75] (q2) at (0.410033, 2.0689) {\rotatebox{-90}{\large2}};
        \node[fill=Orange, circle, scale=0.75] (q3) at (7.85585, 9.71485) {\rotatebox{-90}{\large2}};
        \node[fill=Cyan, circle, scale=0.75] (q4) at (4.48226, 8.02311) {\rotatebox{-90}{\large1}};
        \node[fill=Cyan, circle, scale=0.75] (q5) at (5.3225, 4.57663) {\rotatebox{-90}{\large3}};
        
        \definecolor{darkgray}{rgb}{0.43, 0.5, 0.5}

        \draw [color=darkgray, ->, style=ultra thick] (p6)--(q3);
        \draw [color=darkgray, ->, style=ultra thick] (q3)--(p2);
        \draw [color=darkgray, ->, style=ultra thick] (p2)--(q1);
        \draw [color=darkgray, ->, style=ultra thick] (q1)--(p1);
        \draw [color=darkgray, ->, style=ultra thick] (p1)--(q2);
        \draw [color=darkgray, ->, style=ultra thick] (q2)--(p3);
        \draw [color=darkgray, ->, style=ultra thick] (p3)--(q4);
        \draw [color=darkgray, ->, style=ultra thick] (q4)--(p5);
        \draw [color=darkgray, ->, style=ultra thick] (p5)--(q5);
        \draw [color=darkgray, ->, style=ultra thick] (q5)--(p4);
        
        \draw [color=darkgray, ->, style=ultra thick] (p4) to [out=250,in=340] (q5);
        \draw [color=darkgray, ->, style=ultra thick] (q5) to [out=165,in=270] (p5);
        \draw [color=darkgray, ->, style=ultra thick] (p5) to [out=60,in=295] (q4);
        \draw [color=darkgray, ->, style=ultra thick] (q4) to [out=140,in=30] (p3);
        \draw [color=darkgray, ->, style=ultra thick] (p3) to [out=225,in=100] (q2);
        \draw [color=darkgray, ->, style=ultra thick] (q2) to [out=310,in=210] (p1);
        \draw [color=darkgray, ->, style=ultra thick] (p1) to [out=325,in=225] (q1);
        \draw [color=darkgray, ->, style=ultra thick] (q1) to [out=45,in=285] (p2);
        \draw [color=darkgray, ->, style=ultra thick] (p2) to [out=87,in=307] (q3);
        \draw [color=darkgray, ->, style=ultra thick] (q3) to [out=180,in=75] (p6);
    
    \end{tikzpicture}}
    \end{adjustbox}
     \caption{The first three samples generated by RCS. Notice that the 3 pursuers start $\lceil d/n \rceil = 7$ units apart from each other on $D$.}
  \end{subfigure}
    \caption{The stages of RCS.}
  \label{fig:dfs}
\end{figure*}

\subsection{Method of Sampling}\label{sec:sampling}
Armed with an rSG-PEG suitably enhanced to generate robust solutions, it remains to devise a sampling strategy tailored to place \jpcs in locations likely to lead to a solution quickly.
Because of the computational expense of maintaining the list of non-dominated failure shadow labels at each vertex, it is of paramount importance that each sample we add to the rSG-PEG captures crucial and unique data. 
To do this, we rely on a method of scattering points throughout $F$ called webs. This method was previously introduced by Olsen et al.~\cite{OlsTum+21}, and is enhanced below.  Using webs, we generate a sequence of samples with high connectivity between successive samples and strong coverage of the environment by multiple pursuers.

\subsubsection{Sparse Webs}
A \emph{sparse web}, $W = P \cup Q$, is the union of two sets of points from the environment. The \emph{initial points} $P$ are placed sequentially and uniformly at random outside of the visibility polygon of any previously placed initial points. This process continues until $\bigcup_{p \in P} V(p) = F$. Next, we construct the intersection points $Q$. For each unique pair of points $(p_i, p_j) \in P \times P$, we add to $Q$ a uniformly random point from $V(p_i) \cap V(p_j)$ if $V(p_i) \cap V(p_j) \ne \emptyset$ and $Q \cap V(p_i) \cap V(p_j) = \emptyset$. The final condition, which is the alteration over the original form of webs, reduces the number of points in $W$ while maintaining coverage and connectivity. See Figure~\ref{fig:webs}. The initial points ensure that we have complete visibility coverage of the environment while the intersection points provide straight-line connectivity between the initial points, allowing a pursuer to move freely about an entire sparse web.  Sparse webs generated in this way are used to generate sample \jpcs, as described next.

\subsubsection{Robust cycle samples (RCS)}\label{sec:rcs}
We wish to generate samples in a way that is effective in finding a 1-failure robust solution.
To do this, we construct a graph $H$ whose vertex set is a sparse web $W$ and whose edge set consists of all unique pairs of points from $W \times W$ who can be joined by a line segment contained within the environment.
The construction of webs ensures that $H$ is, in essence, a visibility roadmap of $F$.
Once $H$ is constructed, a random vertex $h$ is selected as the root node for a depth first search (DFS) on $H$.
Throughout the DFS, we construct an ordered list $D$ of vertices visited, both on their initial discovery and via backtracking.
We continue this process until each point of $H$ has been discovered and we successfully backtracked to the root node. We do not add the root node to $D$ during the final stage of the DFS backtracking.
This process produces a list of points $D = (D[0], \ldots, D[d-1])$
which contains every point from $W$ at least once.  The construction also guarantees connectivity between any adjacent points in $D$, giving a spanning walk, with repeats, around $W$.
Since $h$ is the first element added to $D$, $h=D[0]$.

To generate $n$-robot \jpcs, we evenly space the $n$ pursuers along $D$ and walk along the cycle generated by the DFS on $H$.
That is, we create the first sample \jpc by choosing $D[(i-1) \cdot d/n]$ for each robot $i = 1 \dots n$. 
Subsequent sample \jpcs are generated in a similar fashion.  See Figure~\ref{fig:dfs}.  To generate the $k^{\text{th}}$ subsequent sample, we select $D[((i-1) \cdot d/n + k \mod d)]$ for each robot $i =  1 \dots n$ and for each $k = 1, \dots 2d/n$. Having $k$ range over $1, \dots, 2d/n$ ensures that each point in $D$ has been visited by at least two pursuers. Thus, should a failure occur for any single pursuer, that region of the environment has been seen by at least one other pursuer.

\newcommand{\spider}{Figure~\ref{fig:webs}\xspace}
\newcommand{\brick}{Figure~\ref{fig:sols}\xspace}
\newcommand{\nineroom}{Figure~\ref{fig:notation}\xspace}

\newcommand{\tables}{Tables~\ref{tab:3robots}, \ref{tab:4robots} and \ref{tab:5robots}\xspace}
\newcommand{\tablefour}{Table~\ref{tab:4robots}\xspace}
\newcommand{\tablefive}{Table~\ref{tab:5robots}\xspace}
\newcommand{\dfs}{DFS\xspace}
\newcommand{\ws}{WS\xspace}

\section{Evaluation}\label{sec:experiments} 
\begin{figure}[t]
    \centering
    \includegraphics[width=.95\columnwidth, angle=0]{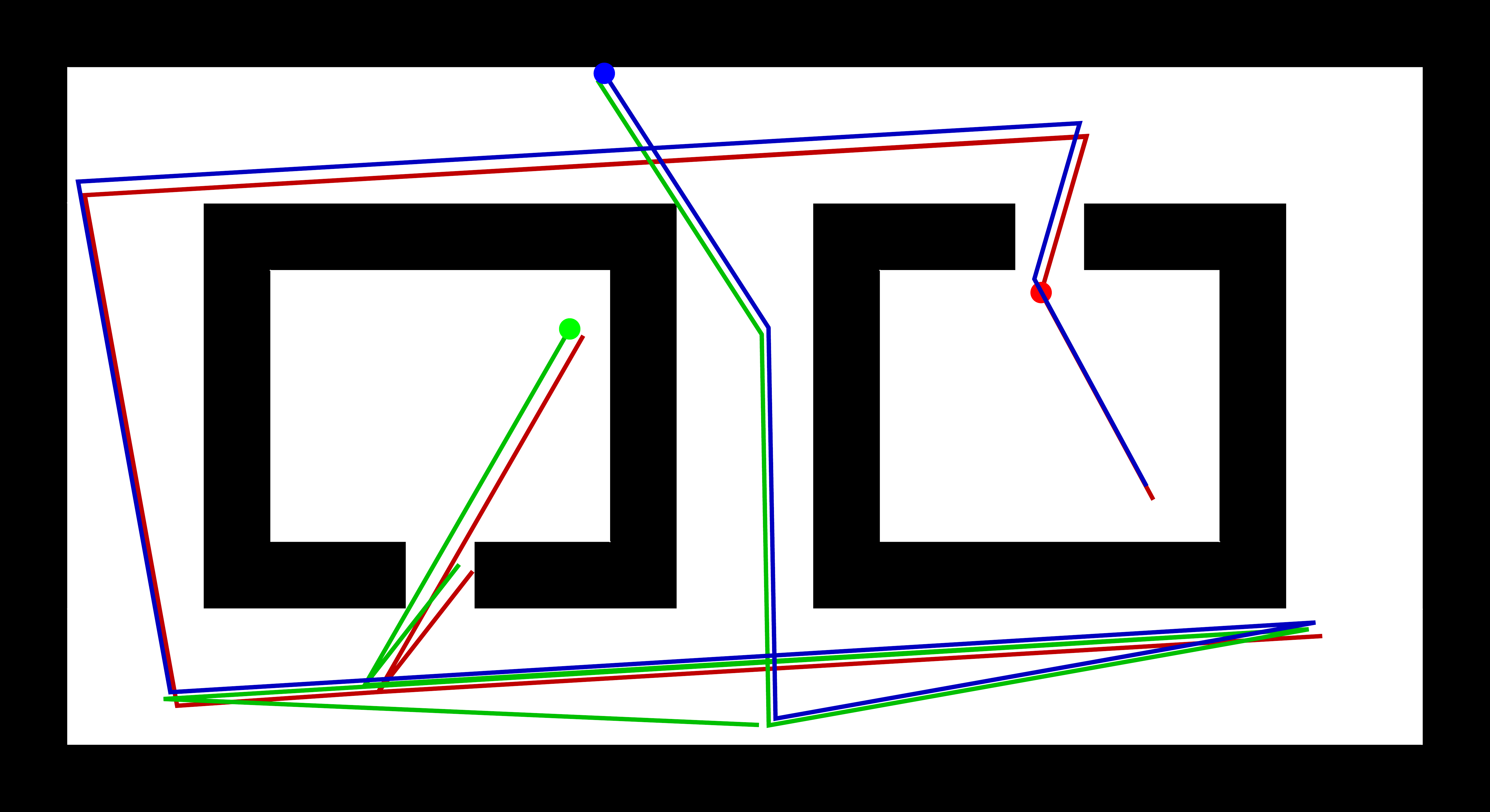}
    \caption{A 1-failure robust solution. Notice each unique region of the environment is examined by at least 2 pursuers. The paths taken by the pursuers were slightly shifted for illustrative purposes.}
  \label{fig:sols}
\end{figure}

\pagebudget{1.25}
We implemented the algorithms described in Section~\ref{sec:ad} in C++. All of the experiments below were conducted on a single core of a 6 core Intel i5-9600K CPU running 64-bit Ubuntu 20.04 at 3.7 GHz with 16 GB of RAM. 
Figure~\ref{fig:sols} shows an example of a 1-failure robust solution computed by this implementation, using both shadow influence caching and RCS.

To evaluate the algorithm quantitatively, we compared (a) the new shadow label updating system described in Section~\ref{sec:cache} against the original naive approach, and (b) the RCS technique described in Section~\ref{sec:rcs} to web sampling (WS).

For (a) we executed the main algorithm, using RCS, to find a 1-failure robust solution utilizing 3 pursuers in the environment shown in \spider. 
We conducted 25 trials using shadow influence caching and 25 trials using the naive method that computes shadow influence anew each time.
All 50 trials successfully found 1-failure robust solutions.
The average computation time (in seconds) utilizing shadow influence caching was 47.02 with a standard deviation of 15.12; without shadow influence caching, the average was 425.88 seconds, with a standard deviation of 512.64.
The results demonstrate an overwhelming benefit to the use of shadow influence caching, which the authors, perhaps counter-intuitively, did not observe in the standard (i.e. 0-failure robust) setting.

For (b), we compared the efficiency of planning with RCS in contrast with planning with WS, a sampling strategy originally presented by Olsen et al.~\cite{OlsTum+21}.
To do so, we attempted to find a 1-failure robust solution utilizing  $n=3$, $4$, and $5$ pursuers across the 3 environments found in Figures~\ref{fig:notation}, \ref{fig:webs} and \ref{fig:sols}. 
For each such scenario, we conducted 50 simulations.
\tables encapsulate the results, showing the mean ($\mu$) and standard deviation ($\sigma$) of the planning time (in seconds) as well as number of vertices and edges in the rSG-PEG. Each such simulation was allotted a timeout of 10 minutes; if the corresponding simulation failed to produce a 1-failure robust solution in that time, it was deemed a failure. 
The results for 3 robots (Table~\ref{tab:3robots}) show that RCS is, at worst, a marginal improvement over \ws in all experiments. The addition of another robot further separated the results of these sampling methods. In particular, the standard deviation of the planning time was significantly decreased, resulting in more consistent run times. 
The highly connective nature of RCS allows us to increase the number of robots without severely suffering from the curse of dimensionality.

\begin{table*}[t]
    \caption{Simulation results ($n$ = 3).}
    \vspace{1.5mm}
    \tiny
    \centering
    \resizebox{0.75\textwidth}{!}{
        \begin{tabular}{lcr@{\hspace{0.75\tabcolsep}}rr@{\hspace{0.75\tabcolsep}}rr@{\hspace{0.75\tabcolsep}}r}
            & success & \multicolumn{2}{c}{planning time (s)} & \multicolumn{2}{c}{Vertices} & \multicolumn{2}{c}{Edges}\\
            & \multicolumn{1}{c}{rate} & \multicolumn{1}{c}{$\mu$} & \multicolumn{1}{c}{$\sigma$} & \multicolumn{1}{c}{$\mu$} & \multicolumn{1}{c}{$\sigma$} & \multicolumn{1}{c}{$\mu$} & \multicolumn{1}{c}{$\sigma$}  \\ \hline

                \multicolumn{1}{c}{\textbf{\nineroom}}\\ \hline
        
                RCS & 100\% & 131.86 & 64.67 & 50.86 & 19.77 & 63.58 & 29.68 \\
            
                WS & 98\% & 151.10 & 50.98 & 59.76 & 16.75 & 73.92 & 26.58 \\
            
                \hline
    
            \multicolumn{1}{c}{\textbf{\spider}}\\ \hline
        
                RCS & 100\% & 50.85 & 20.10 & 23.36 & 7.19 & 30.62 & 11.62 \\
            
                WS & 100\% & 53.01 & 20.22 & 26.56 & 8.14 & 34.44 & 15.04 \\
            
                \hline

            \multicolumn{1}{c}{\textbf{\brick}}\\ \hline
        
                RCS & 100\% & 89.84 & 47.37 & 47.30 & 18.09 & 72.32 & 39.65 \\
            
                WS & 100\% & 109.00 & 57.44 & 50.02 & 18.24 & 77.70 & 41.80 \\
            
                \hline

        \end{tabular}
    }
    \label{tab:3robots}
\end{table*}

\begin{table*}[t]
    \caption{Simulation results ($n$ = 4).}
    \vspace{1.5mm}
    \tiny
    \centering
    \resizebox{0.75\textwidth}{!}{
        \begin{tabular}{lcr@{\hspace{0.75\tabcolsep}}rr@{\hspace{0.75\tabcolsep}}rr@{\hspace{0.75\tabcolsep}}r}
            & success & \multicolumn{2}{c}{planning time (s)} & \multicolumn{2}{c}{Vertices} & \multicolumn{2}{c}{Edges}\\
            & \multicolumn{1}{c}{rate} & \multicolumn{1}{c}{$\mu$} & \multicolumn{1}{c}{$\sigma$} & \multicolumn{1}{c}{$\mu$} & \multicolumn{1}{c}{$\sigma$} & \multicolumn{1}{c}{$\mu$} & \multicolumn{1}{c}{$\sigma$}  \\ \hline
    
            \multicolumn{1}{c}{\textbf{\nineroom}}\\ \hline
        
                RCS & 100\% & 140.57 & 60.01 & 64.72 & 64.57 & 74.34 & 87.02 \\
            
                WS & 100\% & 194.53 & 74.96 & 87.14 & 90.13 & 104.10 & 133.88 \\
            
                \hline

            \multicolumn{1}{c}{\textbf{\spider}}\\ \hline
        
                RCS & 100\% & 58.21 & 21.82 & 23.26 & 24.01 & 28.34 & 40.63 \\
            
                WS & 100\% & 75.40 & 30.11 & 26.56 & 9.21 & 28.54 & 10.95 \\
            
                \hline

            \multicolumn{1}{c}{\textbf{\brick}}\\ \hline
        
                RCS & 100\% & 82.79 & 44.73 & 38.34 & 26.62 & 46.10 & 38.11 \\
            
                WS & 100\% & 113.08 & 57.75 & 46.34 & 25.11 & 52.60 & 33.74 \\
            
                \hline

        \end{tabular}
    }
    \label{tab:4robots}
\end{table*}

\begin{table*}[t]
    \caption{Simulation results ($n$ = 5).}
    \vspace{1.5mm}
    \tiny
    \centering
    \resizebox{0.75\textwidth}{!}{
        \begin{tabular}{lcr@{\hspace{0.75\tabcolsep}}rr@{\hspace{0.75\tabcolsep}}rr@{\hspace{0.75\tabcolsep}}r}
            & success & \multicolumn{2}{c}{planning time (s)} & \multicolumn{2}{c}{Vertices} & \multicolumn{2}{c}{Edges}\\
            & \multicolumn{1}{c}{rate} & \multicolumn{1}{c}{$\mu$} & \multicolumn{1}{c}{$\sigma$} & \multicolumn{1}{c}{$\mu$} & \multicolumn{1}{c}{$\sigma$} & \multicolumn{1}{c}{$\mu$} & \multicolumn{1}{c}{$\sigma$}  \\ \hline
    
            \multicolumn{1}{c}{\textbf{\nineroom}}\\ \hline
        
                RCS & 100\% & 182.10 & 69.91 & 251.26 & 305.28 & 304.30 & 425.75 \\
            
                WS & 98\% & 255.51 & 128.64 & 237.27 & 605.98 & 367.49 & 1233.18 \\
            
                \hline
    
            \multicolumn{1}{c}{\textbf{\spider}}\\ \hline
        
                RCS & 100\% & 84.93 & 46.79 & 88.48 & 238.54 & 258.76 & 1102.28 \\
            
                WS & 100\% & 96.86 & 42.19 & 57.34 & 87.28 & 80.12 & 162.37 \\
            
                \hline

            \multicolumn{1}{c}{\textbf{\brick}}\\ \hline
        
                RCS & 100\% & 86.71 & 47.72 & 65.56 & 95.32 & 90.30 & 183.12 \\
            
                WS & 100\% & 129.01 & 70.41 & 78.50 & 95.09 & 93.82 & 149.29 \\
            
                \hline

        \end{tabular}
    }
    \label{tab:5robots}
\end{table*}

\section{Conclusion}\label{sec:conclusion} \pagebudget{0.25}
This paper addressed the issue of potential single robot failures in the multi-robot visibility-based pursuit-evasion problem. We introduced a modification to an existing data structure to ensure that each solution generated by our algorithm remains a solution in the event of any single robotic failure. The inclusion of shadow caching proved to be cardinal to combat the rapid data expansion in the rSG-PEG. A new sampling method ensured that each region of the environment was observed by two or more pursuers, which allowed us to effectively spread out sampling points and reduce the computation time compared to previous sampling methods. 

Future work may include the extension to $k$-failure robust problems, in the case where $k>1$. In the interest of computation time ---note that the most obvious extension of the present work would require shadow labels of size exponential in $k$--- this general problem would likely require a new data structure which considers some sort of subset lattice structure where the elements represent sets of robots that failed.  

\include{temp-tables2}

\newrefcontext[sorting=nyt]
\printbibliography

\end{document}